\documentclass[letterpaper, 10 pt, conference]{ieeeconf}  

\IEEEoverridecommandlockouts                              
\usepackage{graphics} 
\usepackage{epsfig} 
\usepackage{mathptmx} 
\usepackage{times} 
\usepackage{amsmath} 
\usepackage{amssymb}  
\usepackage{tcolorbox}

\usepackage{booktabs}
\usepackage{varwidth}
\usepackage{algorithm}
\usepackage{multirow}
\usepackage{xcolor}
\usepackage{algpseudocode}
\usepackage{tabularx}
\usepackage[export]{adjustbox}
\algrenewcommand\algorithmicrequire{\textbf{Input:}}
\algrenewcommand\algorithmicensure{\textbf{Output:}}
\algrenewcommand\Require{\item[\kern1.7em\algorithmicrequire]}
\algrenewcommand\Ensure{\item[\kern1.7em\algorithmicensure]}

\newtheorem{dfn}{Definition}
\newtheorem{remark}{Remark}

\DeclareMathAlphabet{\mathcal}{OMS}{cmsy}{m}{n}

\title{\LARGE \bf
Efficient Bayes-Adaptive Reinforcement Learning\\ with Temporal Logic Specifications
}

\author{Jonathan Hau and Alessandro Abate
\thanks{Both authors are with the Department of Computer Science, University of Oxford, United Kingdom
        {\tt\small \{jonathan.hau,alessandro.abate\}@cs.ox.ac.uk}}%
\thanks{\copyright~2026 IEEE. Personal use of this material is permitted.
Permission from IEEE must be obtained for all other uses, in any current or
future media, including reprinting/republishing this material for advertising
or promotional purposes, creating new collective works, for resale or
redistribution to servers or lists, or reuse of any copyrighted component of
this work in other works.}
}

\begin{document}

\maketitle
\thispagestyle{empty}
\pagestyle{empty}

\begin{abstract}

We present a novel end-to-end model-based Reinforcement Learning (RL) algorithm for efficient policy synthesis under given Linear Temporal Logic (LTL) specifications (e.g., safety or reachability) in unknown environments. To do so, a Limit-Deterministic B{\"u}chi Automaton (LDBA) representation of the LTL task is synchronised with a Bayes-Adaptive Markov Decision Process (BAMDP) representation of the environment, which allows us to leverage an enhanced exploration-exploitation trade-off that is achieved via Bayesian RL, as opposed to traditional non-Bayesian approaches. We further propose a novel Bayes-Adaptive Monte-Carlo Planning (BAMCP) algorithm to allow for approximate Bayes-optimal strategy synthesis in the synchronised BAMDP construct. A range of finite- and infinite-horizon task experiments demonstrate the effectiveness of our approach in terms of both property satisfaction and sample efficiency, when compared to traditional model-free approaches. Additional ablation studies also successfully highlight the value of the novel BAMCP algorithm in comparison to classical BAMCP for LTL task satisfaction. Finally, we also showcase a successful application of our approach for \textit{cautious} RL, namely to reduce the number of task violations incurred during policy training.

\end{abstract}

\section{Introduction}
\label{sec:intro}

Linear Temporal Logic (LTL) is an established  formal specification framework that enables the precise definition of non-Markovian, temporal tasks on the state space of a given environment. In this work we construe tasks as goals or objectives, alternatively as requirements or constraints, on the agent. 
As such, LTL is a powerful tool that can be leveraged in Reinforcement Learning (RL) to develop agents that can synthesise policies maximising the probability of satisfying a requirement specified by an LTL formula; this is a framework we refer to as \textit{Logically-Constrained RL (LCRL)} \cite{hasanbeig2019logicallyconstrainedreinforcementlearning}. The majority of the existing work in this area focuses on model-free methods that synthesise policies without learning or maintaining a model of the environment \cite{hasanbeig2019logicallyconstrainedreinforcementlearning,10.1007/978-3-030-17462-0_27,f58b89159342491caff814bc9b4b770c,HASANBEIG2023103949}. Since these approaches work purely through feedback from the environment, they can under-perform in environments with sparse rewards as it is the case of temporally-extended LTL specifications, which provide positive rewards exclusively once fully satisfied (effectively, at the end of the episode). 

Motivated by this observation, we look towards model-based Bayesian RL as an approach that can help to mitigate  such limitations. Through the Bayesian framework, we can represent the MDP as a Bayes-Adaptive Markov Decision Process (BAMDP). This enables an agent to learn by balancing the exploration-exploitation trade-off that is prevalent in environments with unknown dynamics \cite{10.5555/935668,Ghavamzadeh_2015}. In such model-based setups, Monte-Carlo Tree Search (MCTS) is often used as the planning algorithm \cite{deepmindgo,mctspacman}: the Bayes-Adaptive Monte-Carlo Planning (BAMCP) algorithm \cite{guezbamcp} is a known and efficient MCTS planning method for solving BAMDPs. The BAMCP algorithm is an integral part of modern approaches to policy synthesis in unknown environments \cite{pmlr-v205-budd23a,chen2025bayesadaptivemontecarlo,10.5555/3540261.3540349}. As all of these works do not focus on temporally-extended tasks, there is a need to integrate the Bayesian planning approach with the LCRL framework in order to derive an efficient way of learning policies that maximise the satisfaction of LTL objectives. Accordingly, the main contributions of this work are the following:
\begin{enumerate}
    \item We propose a novel, end-to-end model-based RL method that applies a Bayesian planning approach to efficiently synthesise a Bayes-optimal policy for maximising the probability of satisfaction of a given temporal logic specification expressed in LTL. 
    \item To this end, we also propose a planning algorithm that can approximate the Bayes-optimal action at a given hyperstate to \textit{satisfy LTL objectives} within discrete or continuous state spaces.
    \item We perform a wide range of experiments that validate the effectiveness of our approach compared to existing architectures with respect to both the probability of property satisfaction and sample efficiency. Further, we demonstrate the importance of our proposed planning algorithm in Contribution (2) through ablation studies, where our proposed algorithm is shown to clearly outperform classical BAMCP for task satisfaction.
    \item In line with \textit{cautious} RL goals, we further demonstrate the advantages of employing our model-based framework over a model-free approach, by leveraging the model to reduce the number of violations of the specification made during policy training (in addition to the earlier features at deployment). 
\end{enumerate}

\section{Related Work}
\label{sec:relatedwork}

The specification of tasks as LTL formulae in the framework of RL has been studied from many different viewpoints \cite{6426174,degiacomo2019foundationsrestrainingboltsreinforcement,jackermeier2025deepltllearningefficientlysatisfy}, with the majority of approaches focusing on model-free RL \cite{hasanbeig2019logicallyconstrainedreinforcementlearning,10.1007/978-3-030-17462-0_27,f58b89159342491caff814bc9b4b770c} and a few - mostly focussing on safety studies - embracing a model-based setup \cite{hasanbeig2020cautiousreinforcementlearninglogical,mitta2022riskaware}. 
As the latter studies, our Bayes-adaptive approach follows a model-based framework, where the learned generative model of the environment enables leveraging accurate uncertainty quantification to guide and improve exploration in the original context of RL under LTL-specified tasks. 

The work by Voloshin et al. \cite{10.5555/3600270.3601556} applies a model-based approach to satisfying LTL constraints: a constructed model of the environment allows to sample approximately-correct transition dynamics, from which accepting maximal end-components of the underlying MDP can be computed; this reduces policy synthesis to a reachability problem, where an optimal policy can be obtained via Bellman policy iteration on the learnt model. 

Whilst in that work the model is assumed to be fixed in order to obtain the policy downstream, our Bayes-adaptive framework instead learns a Bayesian model alongside the synthesis of the policy that is accordingly optimal for satisfaction of the LTL task. Alternative model-based approaches \cite{COHEN2023101295,10.1145/3447928.3456639} break down the problem into reach-avoid sub-problems, and guarantee specification satisfaction via control barrier certificates: our work instead defines a flexible, LDBA-based reward function with which the synthesised policy automatically guarantees maximal satisfaction probability. Further, our work does not make any restrictive assumptions on the type of MDP or environment.

There exist a few results that have applied the Bayesian framework to control synthesis with LTL specifications:  \cite{wang2021reinforcementlearningtemporallogic} focuses on partially observable MDPs with unknown transition probabilities, using Point-Based Value Iteration \cite{10.5555/1630659.1630806,JMLR:v7:porta06a}, however it does not scale well to high dimensions. In contrast, our work is applicable to all environment types and can easily be adapted for high-dimensional and continuous state and action spaces: here we compute the optimal policy by using a Bayes-Adaptive MDP setup, 
which is similar to some other work \cite{10.5555/3546258.3546547,DBLP:journals/corr/abs-1905-06424}, but these however are standard reward-based approaches that do not work with LTL specifications. 


\section{Preliminaries}
\label{sec:prelims}

\subsection{Markov Decision Processes and RL}
A \textit{Markov Decision Process} (MDP) is a tuple $\langle \mathcal{S}, \mathcal{A}, \mathcal{P}, \mathcal{S}_0, \mathcal{AP}, L \rangle$, where: $\mathcal{S}$ is a set of states; $\mathcal{A}$ is a set of actions; $\mathcal{P}: \mathcal{S} \times \mathcal{A} \rightarrow \mathcal{S}$ is the transition probability function; $\mathcal{S}_0 \subseteq \mathcal{S}$ is the set of initial states; $\mathcal{AP}$ is a finite set of atomic propositions; $L: \mathcal{S} \rightarrow 2^{\mathcal{AP}}$ is a labelling function that assigns a set of atomic propositions to each state.

A (deterministic and memory-less) \textit{policy} $\pi : \mathcal{S} \times \mathcal{A} \rightarrow [0,1]$ is a map from states to a probability distribution over actions. Executing a policy on the MDP generates a (potentially infinite) sequence of states and actions $\tau = s_0a_0s_1a_1...$, called a \textit{trajectory}, such that $s_i \in \mathcal{S}$, $a_i = \pi(s_i)$, and $P(s_i, a_i, s_{i+1}) > 0$ for all $i \geq 0 $. We write $\tau \sim \pi$ to denote that the distribution of possible trajectories depends on the given policy $\pi$.   

Let $ R: \mathcal{S} \times \mathcal{A} \times \mathcal{S} \rightarrow \mathbb{R} $ be a reward function associated with the MDP and $\gamma \in (0,1)$ be the discount factor. Then the goal of RL is to maximise the \textit{expected discounted return}, $J(\pi) = \mathbb{E}_{\tau \sim \pi}[\sum_{t=0}^{\infty}\gamma^t r_t]$, where $r_t = R(s_t, a_t, s_{t+1})$. The \textit{value function} for a given policy is the expected discounted return when starting from state $s$ and executing policy $\pi$, $V^\pi(s) = \mathbb{E}_{\tau \sim \pi}[\sum_{t=0}^{\infty}\gamma^tr_t |s_0 = s]$. The optimal policy $\pi^*$ maximises this value function within the class of policies defined above. 

\subsection{Specifications - LTL and Automata}
We specify temporally-extended requirements via Linear Temporal Logic (LTL). LTL formulae are built up from atomic propositions in $\mathcal{AP}$ according to the following syntax \cite{4567924}: 
\begin{equation}
    \psi ::= true \, | \, a \, | \, \psi \land \psi \, | \, \neg \psi \, | \, \mathbf{X} \psi \, | \, \psi \mathbf{U} \psi. 
\end{equation}
Here $a \in \mathcal{AP}$, and the operators $\mathbf{X}$ and $\mathbf{U}$ are the \textit{next} and \textit{until} operators respectively. 
We further make use of the \textit{always} and \textit{eventually} operators, defined as $\mathbf{G} \psi \equiv \neg \mathbf{F} \neg \psi$, where $\mathbf{F} \psi \equiv true \, \mathbf{U} \psi$.

A given trajectory $\tau$ of an MDP satisfies an atomic proposition $a$ if the initial state is labelled with $a$, i.e., $a \in L(s_0)$. This is denoted as $\tau \models a$. Similarly, $\tau \models \mathbf{X} \psi$ if $\tau[1 \ldots] \models \psi$; finally $\tau \models \psi_1 \mathbf{U} \psi_2$ if $\exists k \geq 0$ s.t. $\tau[k\ldots] \models \psi_2$ and $\forall i < k, \tau[i\ldots] \models \psi_1$. Starting from an initial state $s_0$ of an MDP, we then define the \textit{probability of satisfying an LTL formula} $\psi$ as $\mathbb{P}(\tau^\pi_{s_0} \models \psi)$, where $\tau^\pi_{s_0}$ is the collection of all trajectories $\tau$ generated by the MDP executing policy $\pi$ from the initial state $s_0$, 
and $\mathbb{P}$ denotes the measure on the underlying product probability space (which we implicitly leveraged above to define expected values $\mathbb{E}$).

Any LTL formula can be represented via finite-state automata (e.g., Deterministic Rabin Automata (DRA)  \cite{10.1145/963927.963928} and  Limit-Deterministic B{\"u}chi Automata (LDBA) \cite{Sickertldba, hasanbeig2019logicallyconstrainedreinforcementlearning,Yuan2019ModularDR}). In this work, we utilise LTL-to-LDBA conversion for two reasons: firstly, a DRA representation can result in a doubly exponential blow-up in the size of the automaton \cite{10.1145/963927.963928}, leading to a generally larger product MDP \cite{10.1007/978-3-319-46520-3_9}. Secondly, the LDBA accepting condition is simpler, making it easier to use when constructing a suitable reward function.

An LDBA is a tuple $\langle \mathcal{Q}, q_o, \Sigma, \mathcal{F}, \Delta, \boldsymbol{\varepsilon}  \rangle$ where $\mathcal{Q}$ is a finite set of states; $q_0 \in \mathcal{Q}$ is the initial state; $\Sigma = 2^{\mathcal{AP}}$ is a finite alphabet over the set of atomic propositions; $\mathcal{F}$ is the set of accepting states and $\Delta : \mathcal{Q} \times \Sigma \rightarrow 2^{\mathcal{Q}}$ is the transition function. Further, an LDBA is characterised by the fact that $\mathcal{Q}$ is partitioned into two disjoint sets $\mathcal{Q} = \mathcal{Q}_N \cup \mathcal{Q}_D$ such that $\mathcal{F} \subseteq \mathcal{Q}_D$ and $\Delta(q, \alpha) \in \mathcal{Q}_D$ for all $q \in \mathcal{Q}_D$ and $\alpha \in \Sigma$. In other words, an LDBA is constructed via an initial part $\mathcal{Q}_N$ and an accepting part $\mathcal{Q}_D$; once $\mathcal{Q}_D$ is reached then the automaton cannot leave this set of states. The set $\boldsymbol{\varepsilon}$ contains all the non-deterministic jump transitions, denoted as $\epsilon$-transitions, that enable the automaton to transition from $\mathcal{Q}_N$ to $\mathcal{Q}_D$ without reading any input. An infinite path $\sigma = q_0q_1...$ is accepted by the LDBA if $\mathrm{Inf}(\sigma) \cup \mathcal{F} \neq \emptyset$ where $\mathrm{Inf}$ is the set of states visited by $\sigma$ infinitely often.

\subsection{Synchronising Model and Specification - Product MDP}
\label{ss:product mdp}

The synchronisation of an MDP with an LDBA representation of the LTL specification to be satisfied leads to the construction of a new structure, referred to as the $\textit{product MDP}$: this is useful to employ RL techniques to synthesise policies on the MDP that satisfy the given LTL property. For the setup under study, deterministic and memoryless policies (as defined above) that are function of the product-space, suffice.  

\begin{dfn}[Product MDP]\label{dfn:productmdp}
    For MDP $\mathcal{M} =$ \\ $\langle \mathcal{S}, \mathcal{A}, \mathcal{P}, s_0, \mathcal{AP}, L \rangle$ and LDBA $\mathcal{L} = \langle \mathcal{Q}, q_o, \Sigma, \mathcal{F}, \Delta,  \boldsymbol{\varepsilon}  \rangle$, the \textit{product MDP} is defined as the MDP 
    $$\mathcal{M}^{\psi} = \mathcal{M} \times \mathcal{L} = \langle \mathcal{S}^{\otimes}, \mathcal{A}^{\otimes}, \mathcal{P}^{\otimes}, s_0^{\otimes}, \mathcal{AP}^{\otimes}, L^{\otimes}, \mathcal{F}^{\otimes} \rangle.$$ 
    $\mathcal{M}^{\psi}$ has state space $\mathcal{S}^{\otimes} = \mathcal{S} \times \mathcal{Q}$, action space $\mathcal{A}^{\otimes} = \mathcal{A} \times \boldsymbol{\varepsilon}$, initial state $s_0^{\otimes} = (s_0, q_0)$, set of atomic propositions $\mathcal{AP}^{\otimes} = \mathcal{Q}$, labelling function $\mathcal{L}^{\otimes}: \mathcal{S}^{\otimes} \rightarrow 2^{\mathcal{Q}}$, and the set of accepting states $\mathcal{F}^{\otimes} = \{F_1^{\otimes}, ... \}$ where $F_j^{\otimes} = \mathcal{S} \times F_j$. Finally, if $s_i^{\otimes} = (s_i, q_i)$, then the transition function is given by:
    \begin{equation*}
    \resizebox{\columnwidth}{!}{
        $\mathcal{P}^{\otimes}(s_i^{\otimes}, a, s_j^{\otimes}) = \begin{cases}
            \mathcal{P}(s_i, a, s_j) & \text{if $a \in \mathcal{A}$ and $q_j \in \Delta(q_i, L(s_j)) $, }\\
            1 & \text{if $a = \epsilon_{q_j}$; $q_j \in \Delta(q_i, \epsilon_{q_j})$; $s_i = s_j$, }\\
            0 & \text{otherwise.}
           \end{cases}
           $}
    \end{equation*}
\end{dfn}\vspace{1em}

\section{Problem Definition}
\label{sec:probdef}

We are now equipped with all ingredients to provide the following formal problem statement: 

\begin{tcolorbox}[colback=lightgray!5, colframe=black]
    Considering an unknown MDP $\mathcal{M}$ and a given LTL specification $\psi$ to be satisfied, find an optimal policy $\pi^*$ such that the probability of satisfying the specification from any state is maximised, i.e.: 
    \begin{equation*}
        \pi^* = \underset{\pi}{\mathrm{argmax}} \; \underset{\tau \sim \pi}{\mathbb{E}} \left[ \mathbb{P}(\tau
        \models \psi)  \right]. 
    \end{equation*}
\end{tcolorbox}

\section{Bayes-Adaptive RL with LTL Specifications}
\label{sec:ba-lcrl}

\subsection{Bayesian RL: The Bayes-Adaptive MDP}
\label{ssc:bamdp}

In a Bayesian setting, the agent maintains and updates a \textit{belief} $b$ over the unknown transition and/or reward function, which corresponds to a (posterior) distribution over model parameters that is conditioned on the agent's history of visited states and executed actions $h_t$, i.e., $b_t = p(\mathcal{P}, \mathcal{R}| h_{t})$. The set of possible beliefs $b$ is contained within the \textit{belief space} $\mathcal{B}$. In order to account for model uncertainty during decision making, the state space is augmented with the belief state, resulting in the BAMDP formulation \cite{10.5555/935668,Ghavamzadeh_2015}. Consider an MDP with unknown transition dynamics, $\mathcal{M} = \langle \mathcal{S}, \mathcal{A}, \mathcal{P}, s_0, \mathcal{AP}, L \rangle$, then the corresponding BAMDP is given by the tuple $\langle \mathcal{S}^+, \mathcal{A}, \mathcal{P}^+, s_0^+, \mathcal{AP}, L \rangle$, where: $\mathcal{S}^+ = \mathcal{S} \times \mathcal{B}$ is the set of \textit{hyper-states}, i.e., the Cartesian product of the MDP state space $\mathcal{S}$ and the belief space $\mathcal{B}$ and $s_0^+ \in \mathcal{S}^+$ is the initial hyper-state. $\mathcal{P^+}: \mathcal{S}^+ \times \mathcal{A} \rightarrow \mathcal{S}^+$ is the transition probability function for the BAMDP and is derived as follows:
\begin{align*}
    \mathcal{P^+}(s^+_{t+1} | s^+_t, a_t) &= \mathcal{P^+}(s_{t+1}, b_{t+1} | s_t, a_t, b_t) \\
    &= \mathcal{P^+}(s_{t+1}| s_t, a_t, b_t)\mathcal{P^+}(b_{t+1} | s_t, a_t, b_t, s_{t+1}) \\
    &= \mathbb{E}_{b_t}[\mathcal{P}(s_{t+1}| s_t, a_t)] \delta(b_{t+1} = p(\mathcal{P}| h_{:t+1}) ). 
\end{align*}

It holds that the optimal policy for the BAMDP maximises the following value function in each augmented state \cite{10.5555/935668}: 
\begin{equation}
    V^{*}(s^+_t) = \underset{a}{\mathrm{max}} \sum_{s'} \mathcal{P^+}(s_{t+1} | s^+_t, a_t) \left[r(s, a, s') + \gamma V^{*}(s^+_{t+1})  \right].
\end{equation}

A policy that maximises this value function is called \textit{Bayes-optimal} and optimises decision making under the uncertainty of the unknown transition function.

\subsection{Product BAMDP}
\label{sss: product bamdp}

In Section \ref{ss:product mdp}, we motivated the construction of the $\textit{Product MDP}$ to enable the synthesis of policies on the MDP that also satisfy the given property. In our setup, the dynamics of the environment are unknown and hence a Bayesian treatment allows the natural balance between exploration and exploitation. Hence, we synchronise the BAMDP representation with the LDBA to form the $\textit{Product BAMDP}$. This is defined in an analogous way to the product MDP (see Definition \ref{dfn:productmdp}), but instead the product state space is derived as $\mathcal{S}^{\otimes} = \mathcal{S}^+ \times \mathcal{Q}$ (i.e., the synchronisation of the LTL automata state with the BAMDP hyperstate $s^+ \in \mathcal{S}^+$ rather than with just the MDP state $s \in \mathcal{S})$.

Given that LTL specifications define temporal properties, then a policy satisfying the property may be non-Markovian (i.e., rely on some knowledge of historical states or actions). The product BAMDP allows us to simultaneously keep track of state progression in both the BAMDP and the LDBA. As such, we can consider only Markovian (memoryless) policies of the form $\pi(a | s^+, q)$, as the additional state dimension adds the required memory to track LTL specification satisfaction \cite{10.5555/1373322}. 

\begin{remark}
Importantly, our algorithm tracks the state of the product BAMDP \textit{without explicitly building it} in advance, namely, we only compute the product BAMDP states as they are needed. As such, our algorithm avoids any scalability issues caused by the additional automata state dimension in the product BAMDP state.
\end{remark}

\subsection{Bayes-Adaptive Monte-Carlo Planning in Product-BAMDPs}
\label{ssc:bamcpalg}

The dynamics of a BAMDP (and hence product BAMDP) can be expressed analytically, and hence it is theoretically possible to solve it as a general MDP. However, in practice it is often computationally intractable to do so \cite{guezbamcp}. A popular approximate Bayes-Adaptive algorithm that addresses this issue is the Bayes-Adaptive Monte-Carlo Planning (BAMCP) algorithm \cite{guezbamcp}. This applies Monte-Carlo Tree Search (MCTS) to the BAMDP model to enable an agent to approximately determine the best action at a given hyper-state by using a tree search. However, previous works of BAMCP for both discrete and continuous cases were limited to pure BAMDPs \cite{guezbamcp, chen2025bayesadaptivemontecarlo}. In this section, we present a novel planning method to approximate the Bayes-optimal action at a given hyperstate for \textit{product} BAMDPs with discrete or continuous state spaces.

\paragraph{Overview of the Algorithm} Algorithm \ref{alg:p_bamcp} illustrates the high-level overview of the proposed algorithm, which we refer to as \textit{Product-BAMCP (P-BAMCP)}. As with standard BAMCP, each simulation conducts a tree search from the initial root node to some unvisited node. Action selection is performed using the Upper Confidence Tree (UCT) method \cite{10.1007/11871842_29} in the discrete environments and the Polynomial Upper Confidence Tree (PUCT) method \cite{10.1007/978-3-642-40988-2_13} in the continuous environments, where we have modified the approaches to incorporate the product hyperstates $s^{\otimes}=(s, b, q)$ instead of the standard MDP states. 

After each action selection, we use the sampled model to obtain the successor state $s^{'} \sim \mathcal{P}_{\theta}(s, a)$. The belief vector is updated (we discuss the belief vector later in this section), and we also obtain the successor automata state by extracting the label of the successor state and applying the transition function of the LDBA. A reward from the transition is then obtained from our adaptive reward function $R_{\Phi}: \mathcal{S}^{\otimes} \times \mathcal{A} \times \mathcal{S}^{\otimes} \rightarrow \mathbb{R}$ that we detail in Section \ref{ss:taskmod}. We recursively repeat this process until we reach an unvisited node. An approximation of the value of the unvisited node is obtained and backpropagated through the search tree back to the root node. Repeating this process for $n_{sims}$ allows us to approximate the Bayes-optimal action at the root node in the product BAMDP.

\paragraph{Leaf Node Value Approximation}An unvisited node represents the leaf node of the current search tree, and requires a method to approximate the value function at the leaf node (referred to as \texttt{Value}$(s^{\otimes})$ in Algorithm \ref{alg:p_bamcp}). The exact method depends on the type of environment. In the discrete case, we employ the use of Monte-Carlo estimation where a rollout policy $\pi_{ro}$ is used to select actions for the remainder of the trajectory simulation (until termination or reaching the search horizon). The reward obtained from this rollout is then back-propagated through the nodes in the trajectory. For our experiments, this rollout policy is the currently learned policy obtained as a result of applying Q-learning on the samples $(s,a,r,s')$ generated from interactions between the P-BAMCP agent and the environment \cite{guezbamcp}. In the continuous state case, we directly query the neural network (in this work we employ a Deep Q-Network (DQN) \cite{dqnpaper}).

\paragraph{Sink States} In classical BAMCP, if the search tree reaches the predefined search horizon $d_{max}$ then the search is terminated early and a value of 0 is backpropagated. In P-BAMCP, we track the automata state in order to track the task progression and hence we have another early-stopping condition to consider. 
\begin{dfn}[Non-accepting Sink Component]
    Let $\mathcal{L} = \langle \mathcal{Q}, q_o, \Sigma, \mathcal{F}, \Delta,  \boldsymbol{\varepsilon}  \rangle$ be an LDBA. Then a non-accepting sink component of $\mathcal{L}$ is a subset $Q \subset \mathcal{Q}$ such that the states of $Q$ form a strongly connected graph that does not contain all accepting sets in $\mathcal{F}$, and of which there is no superset of $Q$ that also forms a strongly connected graph. 
\end{dfn}

When a non-accepting sink component is entered, then it is impossible to escape from given the strongly connected nature of the component. Further, as the sink component does not contain all of the accepting states, then it is impossible to visit all of the accepting sets infinitely often. As such, reaching any non-accepting sink component will result in the inability to satisfy the required LTL specification. We denote the union of all non-accepting sink components as the set $\mathfrak{N}$. Accordingly, if the search tree ever enters a state with $q \in \mathfrak{N}$, then we can terminate the search tree early and backpropagate a value of 0 (since we know the task has been failed).

\paragraph{Posterior Distribution} As detailed in Section \ref{ssc:bamdp}, we need to maintain a belief over the unknown parameters and then update them via \texttt{Belief\_Update}$(b)$ after experiencing a new transition. In the discrete case, we leverage a Dirichlet-Multinomial model, a known class of probabilistic models that takes advantage of conjugacy between the two distributions, reducing the Bayesian updates of posterior parameters to observed visit counts of each state \cite{Ghavamzadeh_2015}. In the continuous environments, we can no longer feasibly track counts of all individual transition tuples. Instead, we make use of the fact that an ensemble of neural networks can approximate a function with uncertainty quantification \cite{NIPS2017_9ef2ed4b}. We learn an ensemble of dynamics models $\{\mathcal{P}_{\theta}^1, ..., \mathcal{P}_{\theta}^N \}$, where $\theta$ parametrises the unknown transition function. The belief over the unknown transition function (and/or reward function) can be considered to be a vector of $N$ probabilities, where the $i^{th}$ probability denotes the probability of being in an MDP governed by $\mathcal{P}_{\theta}^i$. Initially, we consider there to be an equal probability for each dynamics model, $b_0 = [1/N, ..., 1/N]$. After observing a transition tuple $(s, a, r, s')$, we can update the belief as follows \cite{chen2025bayesadaptivemontecarlo}:
\begin{equation*}
    b_{t+1}(i) = b_t(i)\mathcal{P}_{\theta}^i(s'|s, a)\mathcal{R}_{\theta}^i(s'|s, a), \hspace{1em} i = 1, ..., N.
\end{equation*}

\begin{algorithm}[tb]
\caption{Product BAMCP (P-BAMCP)}\label{alg:p_bamcp}
\begin{algorithmic}[1]
    \Require $s^{\otimes}, n_{sims}, d_{max}$
    
    \Function{Search}{$s^{\otimes}=(s, b, q), n_{sims}$}
        \For{$n=1...n_{sims}$}
            \State $\theta \sim P(\theta|b)$
            \State \texttt{Simulate}($s^{\otimes}, \theta$)
        \EndFor
        \State \Return $\underset{a}{argmax} \hspace{0.05cm} 
 Q(s^{\otimes}, a)$
    \EndFunction
    \State

    \Function{Simulate}{$s^{\otimes}, \theta, d$}
        \If{$d \geq d_{max}$ or $q \in \mathfrak{N}$}
            \State \Return 0
        \EndIf
        \If{$N(s^{\otimes}) = 0$}
            \For{$a \in \mathcal{A}$}
                \State $N(s^{\otimes}, a) \leftarrow 0$
                \State $Q(s^{\otimes},a) \leftarrow 0$
            \EndFor
            \State $R \leftarrow$ \texttt{Value}$(s^{\otimes})$
            
            \State $N(s^{\otimes}) \leftarrow N(s^{\otimes}) + 1$
            \State $N(s^{\otimes}, a) \leftarrow N(s^{\otimes},a) + 1$
        \Else
            \State $a \leftarrow \underset{j \in \mathcal{A}}{argmax} \hspace{0.05cm}  Q(s^{\otimes}, j) + c P_{UCT}\frac{\sqrt{N(s^{\otimes})}}{1+N(s^{\otimes}, a)}$
            \State $s^{'} \sim \mathcal{P}_{\theta}(s, a)$
            \State $b^{'} \leftarrow$ \texttt{Belief\_Update}$(b)$
            \State $q^{'} \leftarrow \Delta(q, L(s')) $
            \State $r \sim R_{\Phi}(s^{\otimes},a, s^{\otimes'}))$
            \State $N(s^{\otimes}) \leftarrow N(s^{\otimes}) + 1$
            \State $N(s^{\otimes}, a) \leftarrow N(s^{\otimes},a) + 1$

            \State $R \leftarrow r + \gamma \texttt{Simulate}(s^{\otimes'}, \theta, d+1)$ 
        \EndIf

        \State $Q(s^{\otimes},a) \leftarrow Q(s^{\otimes},a) + \alpha[R - Q(s^{\otimes},a)]$
        \State \Return $R$

    \EndFunction

\end{algorithmic}
\end{algorithm}

\section{BA-LCRL Algorithm}
\label{sec:proposedalg}

\subsection{Overview of the Approach}

The pseudocode for BA-LCRL is described in Algorithm \ref{alg:overview}, and consists of four main steps:
\begin{enumerate}
    \item \textbf{LTL-to-LDBA Conversion:} Given any LTL specification, we convert it to an equivalent LDBA representation.
    \item \textbf{P-BAMCP:} At each step, we leverage the learned model and plan using P-BAMCP to determine the next best action to execute in the environment.
    \item \textbf{Bayesian Inference:} The transition tuple $(s, a, r, s')$ generated from executing the action in the environment is used to update our belief.
    \item \textbf{Update Q-Values:} The table/network used to estimate Q-values is updated (e.g., using sampled trajectories from the replay buffer).
\end{enumerate}

\begin{algorithm}[tb]
\caption{Overview of BA-LCRL}\label{alg:overview}
\begin{algorithmic}[1]
    \Require LTL Specification $\psi$
    \Ensure Bayes-optimal policy $\pi^*$

    \State Convert $\psi$ to LDBA $\mathcal{L}$
    \State Initialise frontier sets (accepting $\mathbb{F}$, tracking $\mathbb{T}$) and sink set $\mathfrak{N}$
    \State Initialise $step = 0$ and $epoch = 0$
    \While{not converged}
        \State $epoch \leftarrow epoch + 1$
        \State $s^{\otimes} = (s^+_0, q_0)$
        \While{$(q \notin \mathfrak{N}) \wedge (step < step\_threshold)$ }
            \State $step \leftarrow step + 1$
            \State $a^* \leftarrow \text{P-BAMCP}(s^{\otimes})$ 
            \State Execute $a^*$ and move to $s'$
            \State Update $q \leftarrow q'$ via known automaton transitions  
            \State Update $b \leftarrow b'$ via Bayesian Inference  
            \State Receive reward $R_{\Phi}(s^{\otimes}, a^*, s^{\otimes'})$
            \State Update $\mathbb{F} \leftarrow Acc(q', \mathbb{F})$ and $\mathbb{T} \leftarrow T_{\mathcal{L}}(q', \mathbb{T})$
            \State Update $Q^{\pi}(s^{\otimes}, a^*)$   
            \State $s^{\otimes} \leftarrow s^{\otimes'}$
        \EndWhile
    \EndWhile
\end{algorithmic}
\end{algorithm}

\subsection{Handling Sparse Rewards in LTL-augmented State Spaces: Reward Shaping}
\label{ss:taskmod}
The generation of rewards from LTL objectives in RL is typically achieved via some form of reward machine \cite{10.1613/jair.1.12440}. In this work, we use an LDBA-based adaptive reward function similar to Hasanbeig et al. \cite{hasanbeig2019logicallyconstrainedreinforcementlearning}:

\begin{equation}
    R(s^{\otimes}, a, s^{\otimes'}) = \begin{cases}
            r_p & \text{if $q' \in \mathbb{F}$, } \\
            0 & \text{otherwise,}
           \end{cases}
\end{equation}
where we refer to the set $\mathbb{F}$ as the \textit{accepting frontier set} which tracks the unvisited accepting sets that need to be visited to eventually satisfy the LTL specification.

However, such a reward function only provides rewards upon reaching the accepting sets (i.e., task completion), with no regard to how many steps in the MDP are required before such conditions are achieved. This results in reward sparsity, especially in our Bayes-adaptive setup, as we are applying this reward function over a BAMDP where the state space is infinite (as detailed in Section \ref{ssc:bamdp}).

Following the work by Yuan et al. \cite{Yuan2019ModularDR}, we can augment the base reward function via a potential function $\Phi: \mathcal{S}^{\otimes} \rightarrow \mathbb{R}$ such that the reward function becomes:
\begin{equation}
    R_{\Phi}(s^{\otimes}, a, s^{\otimes'}) = R(s^{\otimes}, a, s^{\otimes'}) + \gamma \Phi(s^{\otimes'}) - \Phi(s^{\otimes}). 
\end{equation}

Importantly, we note that applying such a transformation does not affect the set of optimal policies otherwise obtained using the base reward function \cite{10.5555/645528.657613}, and thus only acts to improve training convergence.

We can then define the following potential function:
\begin{equation}
    \Phi(s^{\otimes}) = \begin{cases}
            \eta r_p & \text{if $q \in \mathbb{T}$, } \\
            0 & \text{otherwise,}
           \end{cases}
\end{equation}
where $0 < \eta < 1 $ is a shaping parameter and the set $\mathbb{T}$ is the \textit{tracking frontier set} which tracks the unvisited intermediate automaton states on the path to an accepting set. Intuitively, this is a reward function that provides a smaller positive reward $ \eta r_p$ for traversing the automaton towards an accepting set, and then provides a larger positive reward $r_p$ for reaching an accepting set (thereby mitigating the issue of reward sparsity).

\subsection{Updating the Tracking Sets}
\label{ss:updatetrack}

As detailed in the previous section (and captured in Line 14 of Algorithm \ref{alg:overview}), we need to maintain and update both the frontier sets to ensure that our adaptive reward only provides reward for \textit{progression} towards the accepting sets (so that the agent doesn't just move back and forth between two intermediate automata states and still receive reward), and then for visiting each of the accepting states in turn (as we need the agent to visit \textit{all} of the accepting sets infinitely often to satisfy the LTL objective). 

Consider an LDBA $\mathcal{L} = \langle \mathcal{Q}, q_o, \Sigma, \mathcal{F}, \Delta,  \boldsymbol{\varepsilon}  \rangle$, where $\mathcal{F} = \{F_1, ..., F_n \} $ is the set of accepting sets, $\mathfrak{N} = \{N_1, ..., N_k \}$ is the set of non-accepting sink components and then let $Q_{sink} = \{q \in \mathcal{Q} | q \in  N_i, \forall i \in \{1, ..., k\}\}$ be the set of sink states. Then we can define a tracking frontier function, $T_{\mathcal{L}} : \mathcal{Q} \times \mathcal{Q} \rightarrow 2^{\mathcal{Q}}$ over the tracking frontier set $\mathbb{T}$ and a Boolean $B$, such that upon a transition from product state $s^{\otimes} = (s^+, q)$ to successor product state $s^{\otimes'} = (s^{+'}, q')$ \cite{9506925}:
    \begin{equation*}
        T_{\mathcal{L}}(q', \mathbb{T}) = \begin{cases}
            \mathbb{T} \backslash q' & \text{if $q \in \mathbb{T}$,}\\
            \mathcal{Q} \backslash (q_0 \cup Q_{sink} \cup q') & \text{if $B = True$,}\\
            \mathbb{T} & \text{otherwise,}
           \end{cases}
    \end{equation*}
where we use the Boolean $B$ to only hold True when all accepting sets have been visited once (i.e., when the accepting frontier set $\mathbb{F}$ becomes empty). As such, the tracking frontier function ensures the set $\mathbb{T}$ maintains the set of unvisited non-accepting and non-sink states in the current round. 

We can also define an accepting frontier function, $A_{\mathcal{L}} : \mathcal{Q} \times \mathcal{F} \rightarrow 2^{\mathcal{Q}}$ over the accepting frontier set $\mathbb{F} \subset \mathcal{F}$ \cite{hasanbeig2019logicallyconstrainedreinforcementlearning}:
    \begin{equation*}
        A_{\mathcal{L}}(q, \mathbb{F}) = \begin{cases}
            \mathbb{F} \backslash F_j & \text{if $q \in F_j$ and $\mathbb{F} \neq \{F_j\} $, }\\
            \mathcal{F} \backslash F_j & \text{if $q \in F_j$ and $\mathbb{F} = \{F_j\}$, }\\
            \mathbb{F} & \text{otherwise.}
           \end{cases}
    \end{equation*}
So if $q \in F_j$, then the function call $A_{\mathcal{L}}(q, \mathbb{F})$ will output a set containing the elements of $\mathbb{F}$ minus the accepting set $F_j$ that has just been visited. In the case where $\mathbb{F} = \{F_j\}$, then we "reset" the set $\mathbb{F}$ to contain all accepting states (i.e., set $\mathbb{F} = \mathcal{F}$) and then remove the recently visited accepting set $F_j$. As such, the agent is guided towards visiting each of the accepting sets infinitely often.

\section{Experiments}
\label{sec:experiments}

In this section, we empirically validate the performance of the BA-LCRL algorithm against the model-free approach. We further showcase the advantage of learning a generative model to enhance safe policy training (cautious RL).

\subsection{Experimental Setup}

\paragraph{Environments}

In the discrete case, we evaluate on a 10 x 10 slippery-grid MDP \cite{guezbamcp, hasanbeig2019logicallyconstrainedreinforcementlearning}. For continuous-state experiments, we employ the Cartpole environment from the OpenAI Gym package \cite{1606.01540}. In all environments, state labels are superimposed on the environment, based on the LTL specification $\psi$.

\paragraph{Tasks}
In the discrete case, we consider several classes of tasks: sequential (non-Markovian) \textbf{reachability} tasks ($\psi_1: \text{F} g_1$ and $\psi_3: \text{F} (g_1 \wedge \text{XF} (g_2 \wedge \text{XF} g_3))$); \textbf{reach-avoid} tasks ($\psi_2: \text{F} (g_1 \wedge \text{XF} (g_2)) \wedge \text{G} \neg a$); \textbf{infinite-horizon recurrence} requirements ($\psi_4: \text{GF} g_1 \wedge \text{GF} g_2$). In the Cartpole environment, we consider two tasks corresponding to keeping the pole upwards whilst avoiding unsafe zones ($\psi_5: \text{G} up \wedge \text{G} \neg a$) and an extension to also traverse the cart to a given position ($\psi_6: \text{G} up \wedge \text{GF} g_1 \wedge \text{G} \neg a$).

\paragraph{Baselines}
With focus on RL with temporally-extended (non-Markovian) tasks, 
the \textit{LCRL} approach by Hasanbeig et. al. \cite{HASANBEIG2023103949} consists of the model-free, non-Bayesian version of our approach. This baseline tells us whether our approach is able to learn the optimal solution, and also highlights any advantages of employing a model-based setup. 

\paragraph{Performance Metrics}
For finite-horizon tasks ($\psi_1, \psi_2, \psi_3, \psi_6$), we measure the average \textbf{Property Satisfaction Probability} (PSP) which measures the average probability that the given LTL specification was satisfied during the test phase. In line with similar works, performance for infinite-horizon tasks ($\psi_4, \psi_5$) is measured by the average \textbf{number of visits to accepting sets} (i.e., the average number of completed cycles) \cite{jackermeier2025deepltllearningefficientlysatisfy}. For both metrics, a higher value is better.

\section{Results and Discussion}
\label{sec:resanddis}

\subsection{Performance and Competitiveness}

Table \ref{table:sumres} compares the average performance metric achieved during test time between BA-LCRL and the standard LCRL approach. Our BA-LCRL approach clearly outperforms the LCRL baseline in terms of both convergence speed and final satisfaction probability (where there exists a difference). This illustrates that our approach is able to successfully learn and exploit a generative model on-the-fly to achieve more efficient learning without compromising the final satisfaction probability. We note that achieving the maximum score (i.e., reaching the step limit of 200 for $\psi_5$ and achieving 100\% PSP for $\psi_6$) is possible in Cartpole as the dynamics are fully deterministic, unlike in Slippery Grid where there are stochastic transition dynamics.

\begin{table}[b]
\caption{Average performance metric results for each task. Best values are in  bold.}
\resizebox{\columnwidth}{!}{%
\begin{tabular}{@{}cccccc@{}}
\toprule
$\psi$                    & \begin{tabular}[c]{@{}c@{}}MDP\\ $(|\mathcal{S}|, |\mathcal{A}|)$\end{tabular} & Algorithm & Performance Metric & Convergence Ep.   \\ \midrule
\multirow{2}{*}{$\psi_1$} & \multirow{2}{*}{$100, 4$}                                        & BA-LCRL   & \textbf{99.33  $\pm$  0.58}                  & 200               \\
                          &                                                            & LCRL      & 98.11 $\pm$  2.65                  & 200               &                      \\ \midrule
\multirow{2}{*}{$\psi_2$} & \multirow{2}{*}{$100, 4$}                                        & BA-LCRL   & \textbf{99.41  $\pm$  0.28}                  & \textbf{75}                \\
                          &                                                            & LCRL      & 96.78 $\pm$  1.42                  & 125               &                      \\ \midrule
\multirow{2}{*}{$\psi_3$} & \multirow{2}{*}{$100, 4$}                                        & BA-LCRL   & \textbf{99.81  $\pm$  0.17}                  & \textbf{75}                \\
                          &                                                            & LCRL      & 99.50 $\pm$  0.44                  & 125               &                      \\ \midrule
\multirow{2}{*}{$\psi_4$} & \multirow{2}{*}{$100, 4$}                                        & BA-LCRL   & \textbf{4.54  $\pm$  0.54}                  & \textbf{300}                \\
                          &                                                            & LCRL      & 3.94 $\pm$  0.63                  & 400               &                      \\ \midrule
\multirow{2}{*}{$\psi_5$} & \multirow{2}{*}{$\infty, 2$}                                        & BA-LCRL   & 200 $\pm$  0.00                  & \textbf{180}            \\
                          &                                                            & LCRL      & 200 $\pm$  0.00                  & 250               &                      \\ \midrule
\multirow{2}{*}{$\psi_6$} & \multirow{2}{*}{$\infty, 2$}                                        & BA-LCRL   & 100 $\pm$  0.00                  & \textbf{160}             \\
                          &                                                            & LCRL      & 100 $\pm$  0.00                  & 220               &                      \\ \bottomrule
\end{tabular}
}\label{table:sumres}
\end{table}

\subsection{Ablation Study: BAMCP vs. P-BAMCP}

In Section \ref{ssc:bamcpalg}, we introduced P-BAMCP, an algorithm that effectively plans in the product BAMDP. We conducted an ablation study to demonstrate the performance of this algorithm in comparison to the standard BAMCP algorithm. In particular, we trained an agent to satisfy $\psi_3$ and $\psi_4$ using standard BAMCP for the same number of episodes that P-BAMCP required to converge (as per Table \ref{table:sumres}). Table \ref{table:sumnoproduct} summarises the average performance metric achieved by the standard BAMCP algorithm (with the relevant values for P-BAMCP drawn from Table \ref{table:sumres} for reference). In both cases, it is clear that the proposed P-BAMCP algorithm vastly outperforms the classical BAMCP algorithm, with the classical algorithm especially struggling with the infinite horizon specification $\psi_4$ where it only achieved a few random successes. These results demonstrate the importance of the P-BAMCP algorithm when trying to approximate the Bayes-optimal action for satisfying LTL tasks and highlight the necessity of augmenting the search tree states with the automata states to enable the agent to track task progression during the planning phase as well.

\begin{table}[b]
\centering
\caption{Average performance metric results for each task when using classical BAMCP. P-BAMCP values extracted from Table \ref{table:sumres} for reference. Best values are in  bold.}
\begin{tabularx}{0.6\columnwidth}{c >{\centering\arraybackslash}X c >{\centering\arraybackslash}X c >{\centering\arraybackslash}X}
\toprule
$\psi$                    &  Algorithm & Performance Metric   \\ \midrule
\multirow{2}{*}{$\psi_3$} &  P-BAMCP   & \textbf{99.81  $\pm$  0.17}                         \\
                                                                               & BAMCP      & 7.84 $\pm$  2.87                                 \\ \midrule
\multirow{2}{*}{$\psi_4$} &  P-BAMCP   & \textbf{4.54  $\pm$  0.54}                        \\
                                                                               & BAMCP      & 0.08 $\pm$  0.02                                \\ \bottomrule
\end{tabularx}
\label{table:sumnoproduct}
\end{table}

\subsection{Ablation Study: Effect of Reward Shaping}

We conducted an ablation study to validate the necessity of reward shaping (particularly for sequential tasks, as detailed in Section \ref{ss:taskmod}) by performing the multi-step sequential task $\psi_3$ without reward augmentation. With reward shaping, BA-LCRL converged in 75 training episodes. However, without reward shaping, BA-LCRL only achieves an average PSP of 5.67\% in the same number of training episodes (as seen in Table \ref{table:sumabs}). BA-LCRL without reward shaping does eventually converge to a similar maximum PSP, but expectedly requires a lot more training episodes to do so. This highlights the necessity of reward shaping to mitigate convergence issues caused by reward sparsity and guide the agent through the sub-tasks of the LDBA to reach the accepting sets. This is especially important for long sequential tasks consisting of many intermediate automata states in sequence (as seen above, performance in even $\psi_3$ which consists of only 3 sequential sub-tasks is massively impacted without the reward shaping).

\begin{table}[t]
\caption{Average satisfaction probability for task $\psi_3$ using BA-LCRL without additional reward shaping.}
\begin{tabularx}{\columnwidth}{c >{\centering\arraybackslash}X c >{\centering\arraybackslash}X c >{\centering\arraybackslash}X}
\toprule
$\psi$                    & \begin{tabular}[c]{@{}c@{}}MDP\\ ($|S|, |A|$)\end{tabular} & \begin{tabular}[c]{@{}c@{}}PSP at prev. \\ Convergence Ep.\end{tabular} & \begin{tabular}[c]{@{}c@{}}Convergence Ep. \\ w/o Reward Shaping\end{tabular} \\ \midrule
\multirow{1}{*}{$\psi_3$} & \multirow{1}{*}{$100, 4$}                                     & 5.67 $\pm$  2.52                     & 175                                                                      \\ \bottomrule
\end{tabularx}
\label{table:sumabs}
\end{table}

\subsection{BA-LCRL Architecture for Cautious RL}
\label{ss:reduceviolations}

We now showcase an example of the benefit from being able to use a Bayesian planning approach to LCRL to encourage certain behaviours without needing to manipulate the properties or dynamics of the real environment. Specifically, we introduced a one-step lookahead component after the MCTS stage (we will now refer to this approach as \textit{Cautious BA-LCRL}). Let $p_{unsafe}$ be the probability that executing some action $a$ in MDP state $s$ results in a transition to a successor MDP state that contains an unsafe label (i.e., would result in the automata moving into a sink component), and introduce some threshold probability $\tau_{threshold}$. Then, the one step lookahead can be summarised as follows:
\begin{enumerate}
    \item At a given state, perform MCTS to determine the next action to be taken (as per Algorithm \ref{alg:overview}).
    \item Perform a one-step lookahead: Use the model to see if $p_{unsafe} > \tau_{threshold}$. If so, then we instead select the second-best action as determined by the MCTS in (1).
\end{enumerate}

The results in Table \ref{table:comparevio} demonstrate that even just by using a one-step lookahead, we can already noticeably reduce the number of violations incurred during training (at convergence with 40 episodes of training, Cautious BA-LCRL reduces violations by approximately 13\%). As such, our approach showcases strong potential as a framework from which further avenues of model-based safe LCRL research can stem from.

\addtolength{\textheight}{-8em}

\begin{table}[t]
\caption{Comparison between Cautious BA-LCRL and BA-LCRL when training to satisfy $\psi_2$. Best values are in bold.}
\resizebox{\columnwidth}{!}{%
\begin{tabular}{@{}cccc@{}}
\toprule
\multirow{2}{*}{Training Eps} & \multicolumn{2}{c|}{Average Violations} &  \multicolumn{1}{c}{Average PSP}                                                      \\ \cmidrule(l){2-4} 
                              & Cautious BA-LCRL           & \multicolumn{1}{c|}{BA-LCRL}                 & Cautious BA-LCRL              \\ \midrule
0                             & 0                          & \multicolumn{1}{c|}{0}                       & 0                 \\
10                            & 8.23 $\pm$  1.47           & \multicolumn{1}{c|}{\textbf{8.14 $\pm$  1.78}} & 33.88 $\pm$  13.11  \\
20                            & \textbf{11.37 $\pm$  1.81} & \multicolumn{1}{c|}{14.26 $\pm$  1.67}       & 97.67 $\pm$  0.58  \\
30                            & \textbf{16.33 $\pm$  1.53} & \multicolumn{1}{c|}{19.34 $\pm$  1.25}       & 98.33 $\pm$  0.55  \\
40                            & \textbf{18.67 $\pm$  2.57} & \multicolumn{1}{c|}{21.51 $\pm$  2.65}       & 99.35 $\pm$  1.15 \\ \bottomrule
\end{tabular}%
}\label{table:comparevio}
\end{table}

\section{Conclusions}
\label{sec:conc}

In this work we have introduced \textit{BA-LCRL}, a Bayesian model-based approach for synthesising optimal policies for satisfying a task defined through LTL formulae. By leveraging P-BAMCP, our method demonstrates strong performance in terms of satisfaction probability and improved sample efficiency over the traditional model-free approach across a range of finite- and infinite-horizon specifications. Finally, we showcased the potential of our approach by adding a simple one-step lookahead component to the planning stage which reduced the number of task violations incurred during the training phase.

\bibliography{ieeeexample}
\bibliographystyle{IEEEtran}

\end{document}